\documentclass[10pt,twocolumn,letterpaper]{article}

\usepackage{iccv}              

\usepackage{booktabs}

\definecolor{iccvblue}{rgb}{0.21,0.49,0.74}
\usepackage[pagebackref,breaklinks,colorlinks,allcolors=iccvblue]{hyperref}

\def\paperID{2356} 
\def\confName{ICCV}
\def\confYear{2025}

\title{Steerable Pyramid Weighted Loss: Multi-Scale Adaptive Weighting for Semantic Segmentation}

\author{Renhao Lu\\
Meinig School of Biomedical Engineering\\
Cornell University\\
Ithaca, NY, USA\\
{\tt\small rl839@cornell.edu}
}

\begin{document}
\maketitle
\begin{abstract}
Semantic segmentation is a core task in computer vision with applications in biomedical imaging, remote sensing, and autonomous driving. While standard loss functions such as cross-entropy and Dice loss perform well in general cases, they often struggle with fine structures, particularly in tasks involving thin structures or closely packed objects. Various weight map-based loss functions have been proposed to address this issue by assigning higher loss weights to pixels prone to misclassification. However, these methods typically rely on precomputed or runtime-generated weight maps based on distance transforms, which impose significant computational costs and fail to adapt to evolving network predictions. In this paper, we propose a novel steerable pyramid-based weighted (SPW) loss function that efficiently generates adaptive weight maps. Unlike traditional boundary-aware losses that depend on static or iteratively updated distance maps, our method leverages steerable pyramids to dynamically emphasize regions across multiple frequency bands (capturing features at different scales) while maintaining computational efficiency. Additionally, by incorporating network predictions into the weight computation, our approach enables adaptive refinement during training. We evaluate our method on the SNEMI3D, GlaS, and DRIVE datasets, benchmarking it against 11 state-of-the-art loss functions. Our results demonstrate that the proposed SPW loss function achieves superior pixel precision and segmentation accuracy with minimal computational overhead. This work provides an effective and efficient solution for improving semantic segmentation, particularly for applications requiring multiscale feature representation. The code is avaiable at \href{https://anonymous.4open.science/r/SPW-0884}{https://anonymous.4open.science/r/SPW-0884}
\end{abstract}    
\section{Introduction}
\label{sec:intro}
Semantic segmentation involves partitioning an image into meaningful regions based on object classes, enabling a detailed understanding of the scene. It plays a crucial role in various applications, such as tumor identification in medical imaging \cite{gul2022deep}, land-use analysis in satellite imagery \cite{bagwari2023comprehensive}, and safety enhancement in autonomous vehicles \cite{kabir2025terrain}. Standard loss functions, such as cross-entropy and Dice loss, are widely used due to their simplicity and effectiveness in general cases \cite{milletari2016v}. However, these functions often struggle with fine structures, particularly thin objects (e.g., blood vessels) and closely packed regions (e.g., glandular tissues), resulting in suboptimal segmentation accuracy. This challenge is particularly pronounced in scenarios requiring precise boundary delineation, such as medical diagnostics and high-resolution imaging \cite{bokhovkin2019boundary}.

To mitigate these limitations, various weight map-based loss functions have been proposed. These methods assign higher loss weights to pixels that are more prone to misclassification, encouraging the model to focus on challenging regions \cite{ronneberger2015u, liu2022boundary, liu2024enhancing}. Typically, weight maps are generated using distance transforms, which calculate the distance of each pixel to the nearest boundary, placing greater emphasis on edge pixels. For example, pixels near class boundaries receive higher weights since they are more susceptible to errors. However, these approaches come with significant computational costs, especially when weight maps are precomputed or generated dynamically during training \cite{guerrero2018multiclass,zhu2022compound}. Furthermore, traditional weight map-based methods are static, failing to adapt to the evolving predictions of the neural network over time. This lack of adaptability can hinder performance, particularly in dynamic learning environments where model predictions continually refine.

In this work, we introduce a novel Steerable Pyramid-based Weighted (SPW) loss function, designed to address the limitations of existing weight map-based methods. Our approach leverages steerable pyramids, a multi-scale and multi-orientation image decomposition technique, to efficiently generate adaptive weight maps based on both the ground truth and network predictions. First introduced in \cite{portilla2000parametric}, steerable pyramids perform a polar-separable decomposition in the frequency domain, enabling independent representation of scale and orientation. This allows for the extraction of features at different levels of detail, making them particularly well-suited for segmentation tasks requiring multi-scale analysis. The SPW loss function utilizes this decomposition to dynamically emphasize regions across multiple frequency bands, capturing fine structures while preserving global context. Unlike traditional boundary-aware losses that rely on static distance maps, our method integrates network predictions into the weight computation, allowing adaptive refinement during training. This dynamic adaptation ensures that the loss function continuously focuses on challenging regions, enhancing segmentation accuracy over time. Additionally, our approach is computationally efficient, mitigating the high overhead associated with distance transform-based methods.

The key contributions of this work are as follows.
\begin{itemize}
    \item We introduce a novel SPW loss function that generates weight maps for cross-entropy loss using steerable pyramid decompositions. This approach enhances multi-scale structural feature representation, improving segmentation accuracy, particularly for fine structures.
    \item We evaluate the SPW loss function on three public segmentation datasets: SNEMI3D (neurite segmentation in electron microscopy slices), GlaS (gland segmentation in H\&E-stained histology slides), and DRIVE (retinal vessel segmentation in fundus images). Compared to 11 state-of-the-art (SOTA) loss functions, SPW loss demonstrates superior performance on both pixel-wise and cluster-based metrics while introducing minimal computational overhead.
\end{itemize}

\section{Related Work}
Semantic segmentation relies heavily on the choice of loss function, which directly affects model convergence, robustness to class imbalance, and the ability to segment fine structures. Early works focused on pixel-wise classification losses, but recent advances have introduced region-aware, boundary-sensitive, and topology-aware losses to address their shortcomings. Below, we categorize existing loss functions and highlight their limitations, positioning our proposed Steerable Pyramid-based Weighted (SPW) loss as a solution that efficiently captures multi-scale structural features while maintaining computational efficiency.
\paragraph{Pixel-Level Loss Functions}
The cross-entropy loss is a fundamental choice for semantic segmentation, treating the task as per-pixel classification. It is simple and effective, but by default it weighs all pixels equally and independently, which means it does not account for spatial context – a misclassified pixel far from the true boundary is penalized the same as one on the wrong side of a border \cite{borse2021inverseform}. Since most segmentation tasks has significantly fewer object pixels than background, cross entropy will lead to a bias toward the background to minimize the loss value \cite{azad2023loss}. To address this limitation, weighted cross-entropy is widely used. Class-balanced CE applies higher weight to under-represented classes (e.g. vessels or roads) to avoid bias toward background \cite{long2015fully}. Focal loss improves upon CE by down-weighting easy-to-classify pixels, thereby focusing learning on harder pixels \cite{lin2017focal}. This method is effective for extreme class imbalance, as demonstrated in medical and remote sensing applications. However, focal loss does not explicitly model spatial structures, making it less effective for preserving thin or detailed regions. 

\paragraph{Region-Based Overlap Losses} Pixel-wise losses are prone to over-penalizing small errors, leading to spatially inconsistent predictions. Dice loss \cite{milletari2016v} and Jaccard loss (IoU loss) \cite{duque2021power} mitigate this by optimizing set-based similarity, directly improving overlap with ground truth masks. Tversky loss \cite{salehi2017tversky} generalizes Dice by introducing tunable parameters to balance precision and recall, benefiting scenarios where missing small objects is worse than false positives. While these losses improve segmentation accuracy for minority classes, they remain global metrics and do not differentiate between easy and hard regions. Consequently, they may still struggle with boundary accuracy and fine-grained structures, which motivates the need for more adaptive loss functions.

\paragraph{Boundary-Sensitive and Weighted Losses}
Recognizing the importance of boundary refinement, several loss functions incorporate boundary awareness into training. Weighted cross-entropy loss, first introduced in U-Net \cite{ronneberger2015u}, assigns higher importance to pixels near class boundaries, enhancing edge definition. Adaptive Boundary Weighted (ABW) loss \cite{liu2022boundary} extends this idea by dynamically adjusting loss weights based on distance transforms, leading to better object separation. However, traditional boundary-aware losses rely on static or manually designed weight maps, which fail to adapt to evolving network predictions. Furthermore, computing distance transforms at runtime can be computationally expensive. 

\begin{figure*}
    \centering
    \includegraphics[width=0.9\linewidth]{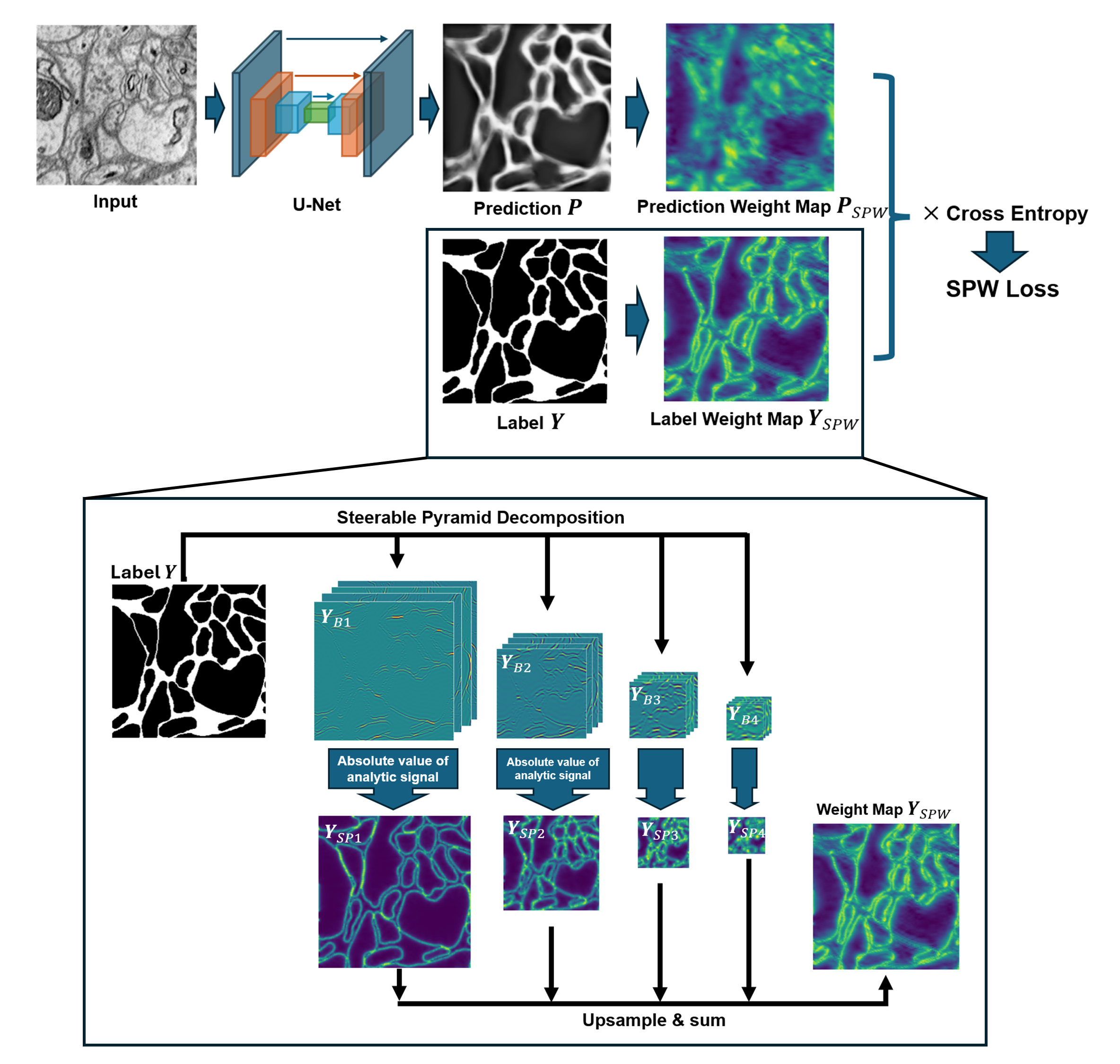}
    \caption{Illustration of the Steerable Pyramid-based Weighted (SPW) Loss. The weighted maps (\(\mathbf{Y}_{SPW}\), \(\mathbf{P}_{SPW}\)) derived from the ground truth label (\(\mathbf{Y}\)) and the predicted probability map (\(\mathbf{P}\)) are incorporated into a weighted cross-entropy loss. To compute \(\mathbf{Y}_{SPW}\), \(\mathbf{Y}\) is first decomposed using a steerable pyramid, producing subbands at multiple scales and orientations (\(\mathbf{Y}_{B_{ik}}\)). The amplitudes of the analytic signal from all subbands are then upsampled and combined to form the final SPW map.}
    \label{fig:1}
\end{figure*}

\paragraph{Structural and Topology-Aware Losses}
Recent advances incorporate structural constraints into loss design, ensuring that segmentation outputs maintain connectivity and spatial coherence. Region Mutual Information (RMI) loss \cite{zhao2019region} models the joint distribution of neighboring pixels, encouraging local consistency in predictions. While this improves fine structure segmentation, it introduces computational overhead due to high-dimensional mutual information calculations. Based on RMI, the complex wavelet mutual information (CWMI) loss were proposed to utilize the complex wavelet decomposition to minimize the structural difference between label and prediction \cite{lu2025complex}. For thin, tubular structures (e.g., vessels, roads), clDice loss \cite{shit2021cldice} ensures connectivity by optimizing a differentiable skeleton-based overlap measure. Skea-Topo loss \cite{liu2024enhancing} extends this by explicitly penalizing topology-breaking errors, preventing incorrect object merges or splits. While effective for preserving structural continuity, these methods do not account for multi-scale feature representation, making them susceptible to errors when objects of varying sizes are present. 
\section{Methods}

For a semantic segmentation task, given the ground truth label \(Y\) and predicted probability map \(P\), the cross entropy loss can be defined as:
\begin{equation}
L_{CE}(Y, P)=\sum_{c}\sum_{\mathbf{x}}Y_{\mathbf{x},c}logP_{\mathbf{x},c}
\end{equation}
where \(c\) is the indexes of classes, and \(\mathbf{x}\) is the coordinates of the pixels. 
However, not all pixels contribute equally to learning—some pixels are harder to classify due to complex structures or boundaries. To emphasize these important regions, we propose the Steerable Pyramid Weighted (SPW) loss (Figure \ref{fig:1}):
\begin{equation}
L_{SPW}=\sum_{c}\sum_{\mathbf{x}}w(\mathbf{x})Y_{\mathbf{x},c}logP_{\mathbf{x},c}
\end{equation}
where the pixel-wise weight function is:
\begin{equation}
w(\mathbf{x})=w_c+\lambda w_{SPW}(Y,P)
\end{equation}
Here, \(w_c\) is the class imbalance weight, and \(w_{SPW}(Y,P)\) leverages steerable pyramid decomposition to assign higher weights to pixels with strong multi-scale textures and boundaries, as described below. \(\lambda\) is a hyperparameter controlling the contribution of SPW weight map.

\subsection{Steerable Pyramid Decomposition}
Accurate segmentation requires detecting edges and textures at multiple scales. A naive approach would be to apply multiple edge detectors at different resolutions, but this is inefficient and lacks theoretical soundness. Instead, we use the steerable pyramid, a structured approach that decomposes an image into multi-scale, multi-orientation components, capturing both fine and coarse structures simultaneously \cite{simoncelli1992shiftable,simoncelli1995steerable}. 

The steerable pyramid works by breaking an image into frequency bands at different orientations, much like a set of specialized edge detectors that operate at different resolutions. This is achieved using a set of band-pass filters that extract image structures at different scales.

In the frequency domain, the band-pass filter for the \(k_{th}\) orientation is expressed in polar coordinates \((r, \theta)\) as:
\begin{equation}
B_k(r, \theta) = H(r) G_k(\theta), \quad k \in [1, K],
\end{equation}
where \(H(r)\) and \(G_k(\theta)\) represent the radial (frequency response) and angular (directional sensitivity) components, respectively:
\begin{equation}
H(r) =
\begin{cases}
    \cos\left(\frac{\pi}{2} \log_2 \left(\frac{2r}{\pi}\right)\right), & \frac{\pi}{4} < r < \frac{\pi}{2}, \\
    1, & r \leq \frac{\pi}{2}, \\
    0, & r \geq \frac{\pi}{4}.
\end{cases}
\end{equation}
\begin{equation}\label{G real}
G_k(\theta) =
    \alpha_k \left|\cos\left(\theta - \frac{\pi k}{K}\right)\right|^{K-1},
\end{equation}
where \(K\) is the number of orientations and \(\alpha_k = 2^{k-1} \frac{(K-1)!}{\sqrt{K[2(K-1)]!}}\).
This decomposition creates a pyramid-like hierarchy, where the image is recursively filtered and downsampled. After \(N\) levels, an image \(\mathbf{I}\) is represented as:
\begin{equation}\label{eq:7}
\mathbf{I} \rightarrow
\left|
\begin{array}{ll}
    \mathbf{I}_{H_0} \in \mathbf{R}^{H_0 \times W_0}, \\
    \mathbf{I}_{B_{11}}, \mathbf{I}_{B_{12}}, \dots, \mathbf{I}_{B_{1K}}  \in \mathbf{R}^{H_0 \times W_0}, \\
    \mathbf{I}_{B_{21}}, \mathbf{I}_{B_{22}}, \dots, \mathbf{I}_{B_{2K}}  \in \mathbf{R}^{H_1 \times W_1}, \\
    \dots \\
    \mathbf{I}_{B_{N1}}, \mathbf{I}_{B_{N2}}, \dots, \mathbf{I}_{B_{NK}} \in \mathbf{R}^{K \times H_{N-1} \times W_{N-1}}, \\
    \mathbf{I}_{L_0} \in \mathbf{R}^{H_{N-1} \times W_{N-1}},
\end{array}
\right.
\end{equation}
where \(\mathbf{I}_{H_0}\) and \(\mathbf{I}_{L_0}\) are high-frequency and low-frequency residues, and \(\mathbf{I}_{B_n}\) represents represents orientation-specific textures at scale \(n\) and \(k\)-th direction. 

\subsection{Weighted Map Using Analytic Signal}
The steerable pyramid decomposition provides detailed multi-scale edge information, but we still need a way to measure the strength of these structures. Image details (e.g., edges and textures) often appear as rapid oscillations in pixel values. Instead of analyzing raw pixel values, we compute the local amplitude envelope, which tells us how strong a texture or boundary is at any location. 

To achieve this, we use the analytic signal, which provides a way to measure local energy in an image \cite{ulrich2006envelope}. The analytic signal of the subband image \(\mathbf{I}\) can be calculated by:
\begin{equation}
    \mathbf{I}_a=\mathbf{I}+j\hat{\mathbf{I}}
\end{equation}
where \(\hat{\mathbf{I}}\) is the Hilbert transform of \(\mathbf{I}\), which shifts phase components by 90 degrees. The benefit of this transformation is that it allows us to compute a smooth amplitude envelope \(|\mathbf{I}_a|\), which eliminates fine oscillations and directly captures edge strength. Mathematically, the analytic signal is equivalent to disregarding the negative frequency in the Fourier domain, which is computationally efficient.

Hence, the weighted map in \(i^{th}\) scale can be defined as the sum of the weight map in each direction (\(k\)):
\begin{equation}
    w_i=\sum_{k=1}^{K}|\mathbf{I}_{B_{ik}}+j\hat{\mathbf{I}}_{B_{ik}}|
\end{equation}
And the weighted map for the entire image is:
\begin{equation}
\label{eq 10}
    w(\mathbf{I})=\sum_{i=1}^{N}\sum_{k=1}^{K}\beta^{i-1}U(|\mathbf{I}_{B_{ik}}+j\hat{\mathbf{I}}_{B_{ik}}|)
\end{equation}
where \(\beta\) is a hyperparameter controlling contributions from different frequency subbands, and \(U(*)\) indicates upsampling by zero padding in the Fourier domain.

\subsection{SPW loss}
Both the label-derived and prediction-derived weight maps contribute to the final SPW map based on equation \ref{eq 10}:
\begin{equation}
\begin{aligned}
    w_{SPW}(Y,P)=&\sum_{i=1}^{N}\sum_{k=1}^{K}\beta^{i-1}U(|\mathbf{Y}_{B_{ik}}+j\hat{\mathbf{Y}}_{B_{ik}}|)+\\
    &\sum_{i=1}^{N}\sum_{k=1}^{K}\beta^{i-1}U(|\mathbf{P}_{B_{ik}}+j\hat{\mathbf{P}}_{B_{ik}}|)
\end{aligned}
\end{equation}
where the label-based weight map emphasizes difficult-to-classify pixels (e.g., edges), while the prediction-based weight map highlights high-confidence, high-frequency features that require verification. All SPW maps are generated on the fly during training without gradient tracking, ensuring minimal overhead. By integrating these two components, the SPW loss naturally focuses learning on challenging regions, improving segmentation accuracy.
\section{Experiment}

\subsection{Experimental Setup}
\paragraph{Datasets}
We tested SPW loss on three public segmentation datasets, all characterized by class and instance imbalance: (1) SNEMI3D, a neurite segmentation dataset containing 100 \(1024 \times 1024\) grayscale images from electron microscopy slices \cite{arganda2013snemi3d}; (2) GlaS, a gland segmentation dataset with 165 RGB images of varying sizes from histological images of colorectal cancer samples \cite{sirinukunwattana2017gland}; and (3) DRIVE, a retinal vessel segmentation dataset comprising 40 \(584 \times 565\) RGB images from fundus photographs \cite{staal2004ridge}; For all datasets, three-fold cross-validation was used to ensure robust evaluation.

\paragraph{Baselines and Implementation Details}
We employed U-Net \cite{ronneberger2015u} as base models to evaluate SPW loss. We compared SPW against 11 state-of-the-art (SOTA) loss functions, including pixel-wise loss functions (e.g., cross entropy, BCE \cite{long2015fully}, Focal loss \cite{ross2017focal}), region-based loss functions (e.g., Dice loss \cite{milletari2016v}, Tversky loss \cite{salehi2017tversky}, Jaccard loss \cite{rahman2016optimizing}), and structural/topological loss functions (e.g., WCE \cite{ronneberger2015u}, ABW loss \cite{liu2022boundary}, Skea-topo loss \cite{liu2024enhancing}, RMI loss \cite{zhao2019region}, clDice loss \cite{shit2021cldice}). Hyperparameters for each baseline were tuned via grid search, with Tversky loss (\(\alpha = 0.5, \beta = 0.5\)) and Focal loss (\(\gamma = 2.5\)) as examples. 

A previous study indicates steerable pyramids with four decomposition levels and four orientations is capable of comparing multiscale features for segmentation tasks \cite{lu2025complex}, hence, we used the same parameter in the steerable pyramid in all experiments. The hyperparameters \(\lambda=10\) and \(\beta=0.9\) were applied in Table \ref{tab:1}, while their determination was revealed in the ablation test section. Adam optimizer with a StepLR scheduler (initial learning rate \(1 \times 10^{-4}\), decay rate 0.8, step size 10) was used, and models were trained for 50 epochs with a batch size of 10. Early stopping based on mIoU was employed to select the best model. Training was conducted on an NVIDIA A100 GPU using the Google Colab runtime.

\paragraph{Data Augmentation and Evaluation Metrics}
Random flips and rotations were applied to all datasets to improve generalization. For SNEMI3D, images were randomly cropped to \(512 \times 512\), while for GlaS, images were cropped to \(448 \times 576\) to standardize input sizes. No cropping was performed for DRIVE due to its uniform image dimensions.

Performance was evaluated using five metrics: mIoU and mDice for regional precision, variation of information (VI) \cite{nunez2013machine} and adjusted Rand index (ARI) \cite{vinh2009information} for clustering precision. These metrics provide a comprehensive assessment of both regional overlap and structural fidelity.

\begin{table}\label{Table 1}
\footnotesize 
    \centering
    \setlength{\tabcolsep}{1pt}
    \begin{tabular}{lcccc}
        \toprule
        \multicolumn{5}{c}{\textbf{SENMI3D}}\\
        \hline
Methods  & 
mIoU$\uparrow$ & mDice$\uparrow$ & VI$\downarrow$ & ARI$\uparrow$ \\

\hline

CE          & \(0.752_{\pm0.010}\) & \(0.850_{\pm0.007}\) & \(1.77_{\pm0.25}\) & \(0.548_{\pm0.033}\) \\
BCE         & \(0.740_{\pm0.006}\) & \(0.843_{\pm0.005}\) & \(1.70_{\pm0.11}\) & \(0.570_{\pm0.001}\) \\
Dice        & \(\underline{0.768_{\pm0.005}}\) & \(\underline{0.862_{\pm0.004}}\) & \(\underline{1.35_{\pm0.06}}\) & \(0.607_{\pm0.015}\) \\
Focal       & \(0.726_{\pm0.005}\) & \(0.834_{\pm0.004}\) & \(1.71_{\pm0.04}\) & \(0.564_{\pm0.011}\) \\
Jaccard     & \(0.765_{\pm0.001}\) & \(0.860_{\pm0.001}\) & \(1.38_{\pm0.04}\) & \(0.608_{\pm0.010}\) \\
Tversky     & \(0.759_{\pm0.006}\) & \(0.857_{\pm0.004}\) & \(1.38_{\pm0.04}\) & \(\underline{0.611_{\pm0.008}}\) \\
WCE         & \(0.715_{\pm0.005}\) & \(0.827_{\pm0.004}\) & \(1.79_{\pm0.02}\) & \(0.547_{\pm0.005}\) \\
ABW         & \(0.671_{\pm0.011}\) & \(0.795_{\pm0.008}\) & \(2.30_{\pm0.11}\) & \(0.465_{\pm0.014}\) \\
Skea-topo   & \(0.667_{\pm0.002}\) & \(0.791_{\pm0.001}\) & \(2.53_{\pm0.18}\) & \(0.446_{\pm0.011}\) \\
RMI         & \(0.768_{\pm0.010}\) & \(0.862_{\pm0.007}\) & \(1.36_{\pm0.15}\) & \(0.599_{\pm0.017}\) \\
clDice      & \(0.712_{\pm0.024}\) & \(0.824_{\pm0.018}\) & \(1.71_{\pm0.13}\) & \(0.548_{\pm0.039}\) \\
\hline
SPW         & \(\mathbf{0.770_{\pm0.003}}\) & \(\mathbf{0.864_{\pm0.002}}\) & \(\mathbf{1.22_{\pm0.06}}\) & \(\mathbf{0.629_{\pm0.010}}\) \\

\toprule
        \multicolumn{5}{c}{\textbf{GlaS}}\\
        \hline
Methods  & 
mIoU$\uparrow$ & mDice$\uparrow$ & VI$\downarrow$ & ARI$\uparrow$ \\

\hline

CE          & \(0.821_{\pm0.012}\) & \(0.897_{\pm0.008}\) & \(0.95_{\pm0.05}\) & \(0.697_{\pm0.020}\) \\
BCE         & \(0.828_{\pm0.013}\) & \(0.902_{\pm0.008}\) & \(0.89_{\pm0.02}\) & \(0.709_{\pm0.028}\) \\
Dice        & \(0.816_{\pm0.011}\) & \(0.894_{\pm0.007}\) & \(0.94_{\pm0.05}\) & \(0.693_{\pm0.024}\) \\
Focal       & \(0.792_{\pm0.003}\) & \(0.878_{\pm0.002}\) & \(1.07_{\pm0.09}\) & \(0.646_{\pm0.016}\) \\
Jaccard     & \(0.829_{\pm0.003}\) & \(0.902_{\pm0.002}\) & \(0.92_{\pm0.03}\) & \(0.713_{\pm0.007}\) \\
Tversky     & \(0.829_{\pm0.008}\) & \(0.902_{\pm0.006}\) & \(0.92_{\pm0.04}\) & \(0.708_{\pm0.005}\) \\
WCE         & \(0.830_{\pm0.016}\) & \(0.902_{\pm0.012}\) & \(0.91_{\pm0.04}\) & \(0.723_{\pm0.023}\) \\
ABW         & \(0.758_{\pm0.020}\) & \(0.856_{\pm0.015}\) & \(1.50_{\pm0.03}\) & \(0.613_{\pm0.023}\) \\
Skea-topo   & \(0.784_{\pm0.008}\) & \(0.874_{\pm0.007}\) & \(1.31_{\pm0.07}\) & \(0.652_{\pm0.010}\) \\
RMI         & \(\underline{0.843_{\pm0.008}}\) & \(\underline{0.911_{\pm0.006}}\) & \(\underline{0.82_{\pm0.02}}\) & \(\underline{0.737_{\pm0.010}}\) \\
clDice      & \(0.810_{\pm0.021}\) & \(0.890_{\pm0.014}\) & \(0.97_{\pm0.10}\) & \(0.696_{\pm0.045}\) \\
\hline
SPW         & \(\mathbf{0.851_{\pm0.005}}\) & \(\mathbf{0.916_{\pm0.003}}\) & \(\mathbf{0.71_{\pm0.04}}\) & \(\mathbf{0.760_{\pm0.001}}\) \\

\toprule
        \multicolumn{5}{c}{\textbf{DRIVE}}\\
        \hline
Methods  & 
mIoU$\uparrow$ & mDice$\uparrow$ & VI$\downarrow$ & ARI$\uparrow$ \\

\hline

CE          & \(0.765_{\pm0.004}\) & \(0.852_{\pm0.004}\) & \(1.42_{\pm0.23}\) & \(0.376_{\pm0.070}\) \\
BCE         & \(0.747_{\pm0.017}\) & \(0.839_{\pm0.014}\) & \(1.39_{\pm0.20}\) & \(0.483_{\pm0.090}\) \\
Dice        & \(0.782_{\pm0.005}\) & \(0.865_{\pm0.004}\) & \(1.33_{\pm0.17}\) & \(0.437_{\pm0.073}\) \\
Focal       & \(0.730_{\pm0.012}\) & \(0.826_{\pm0.010}\) & \(1.39_{\pm0.19}\) & \(0.473_{\pm0.124}\) \\
Jaccard     & \(0.766_{\pm0.007}\) & \(0.854_{\pm0.006}\) & \(\underline{1.24_{\pm0.05}}\) & \(0.526_{\pm0.026}\) \\
Tversky     & \(0.764_{\pm0.006}\) & \(0.853_{\pm0.005}\) & \(1.26_{\pm0.07}\) & \(0.520_{\pm0.032}\) \\
WCE         & \(0.730_{\pm0.012}\) & \(0.826_{\pm0.009}\) & \(1.45_{\pm0.28}\) & \(0.489_{\pm0.082}\) \\
ABW         & \(0.465_{\pm0.008}\) & \(0.506_{\pm0.038}\) & \(1.79_{\pm0.26}\) & \(0.019_{\pm0.028}\) \\
Skea-topo   & \(0.650_{\pm0.040}\) & \(0.750_{\pm0.045}\) & \(1.73_{\pm0.10}\) & \(0.275_{\pm0.120}\) \\
RMI         & \(\underline{0.783_{\pm0.009}}\) & \(\underline{0.866_{\pm0.008}}\) & \(1.32_{\pm0.04}\) & \(0.429_{\pm0.033}\) \\
clDice      & \(0.720_{\pm0.004}\) & \(0.819_{\pm0.004}\) & \(1.30_{\pm0.15}\) & \(\underline{0.550_{\pm0.053}}\) \\
\hline
SPW         & \(\mathbf{0.784_{\pm0.010}}\) & \(\mathbf{0.867_{\pm0.008}}\) & \(\mathbf{1.14_{\pm0.09}}\) & \(\mathbf{0.565_{\pm0.024}}\) \\

\toprule
    \end{tabular}
    \caption{Quantitative results of different loss functions across the three datasets. The \textbf{bold} numbers indicate the best performance for each metric, while the \underline{underlined} numbers denote the second-best performance.}
    \label{tab:1}
\end{table}

\subsection{Quantitative and qualitative results}
\begin{figure}[!h]
    \centering
    \includegraphics[width=\linewidth]{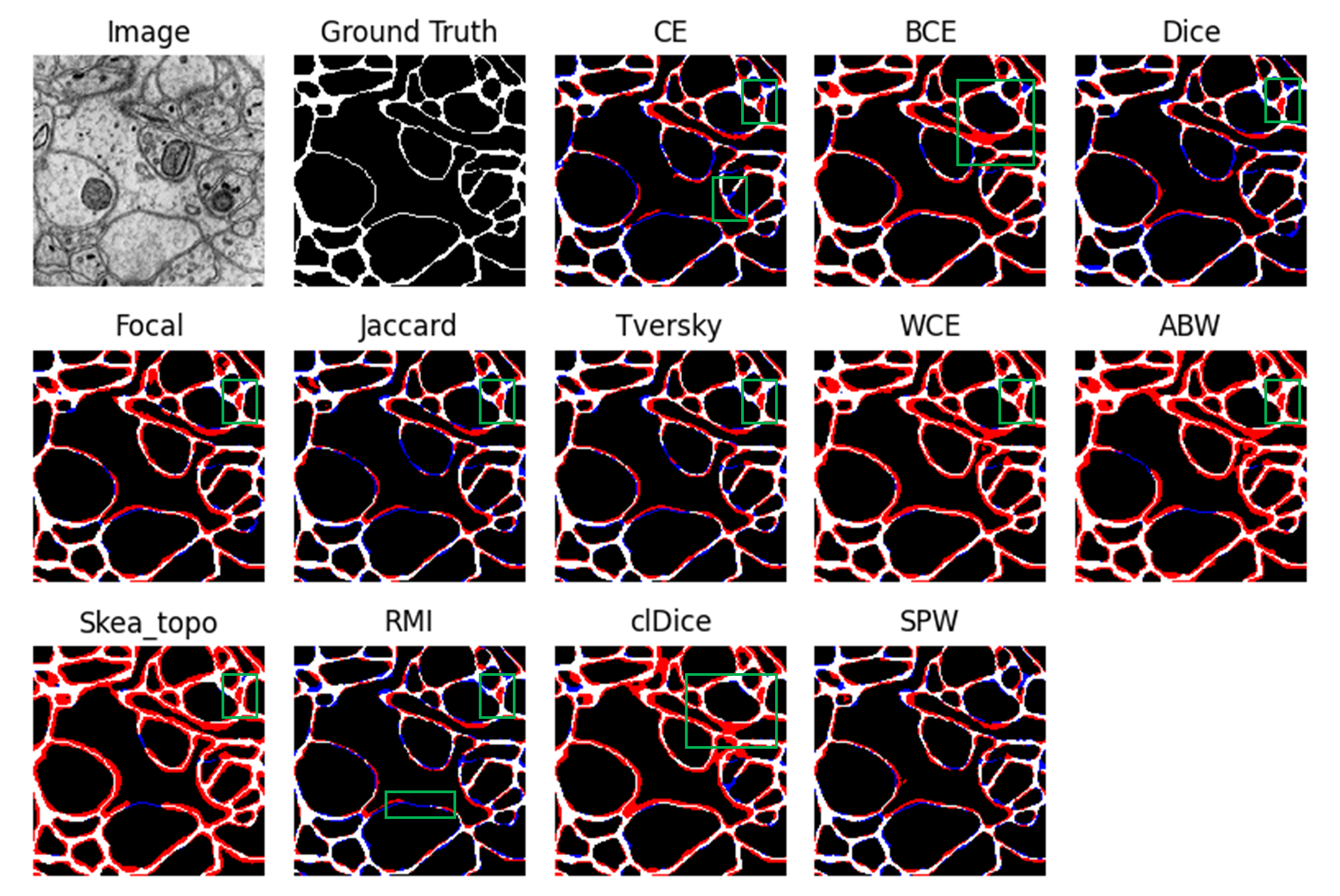}
    \caption{Qualitative results of different loss functions on the SNEMI3D dataset. \textcolor{red}{Red}: false positive regions; \textcolor{blue}{Blue}: false negative regions. \textcolor{ForestGreen}{Green box}: challenging segmentation errors that are successfully addressed by SPW loss.}
    \label{fig:2}
\end{figure}

\begin{figure}[!h]
    \centering
    \includegraphics[width=\linewidth]{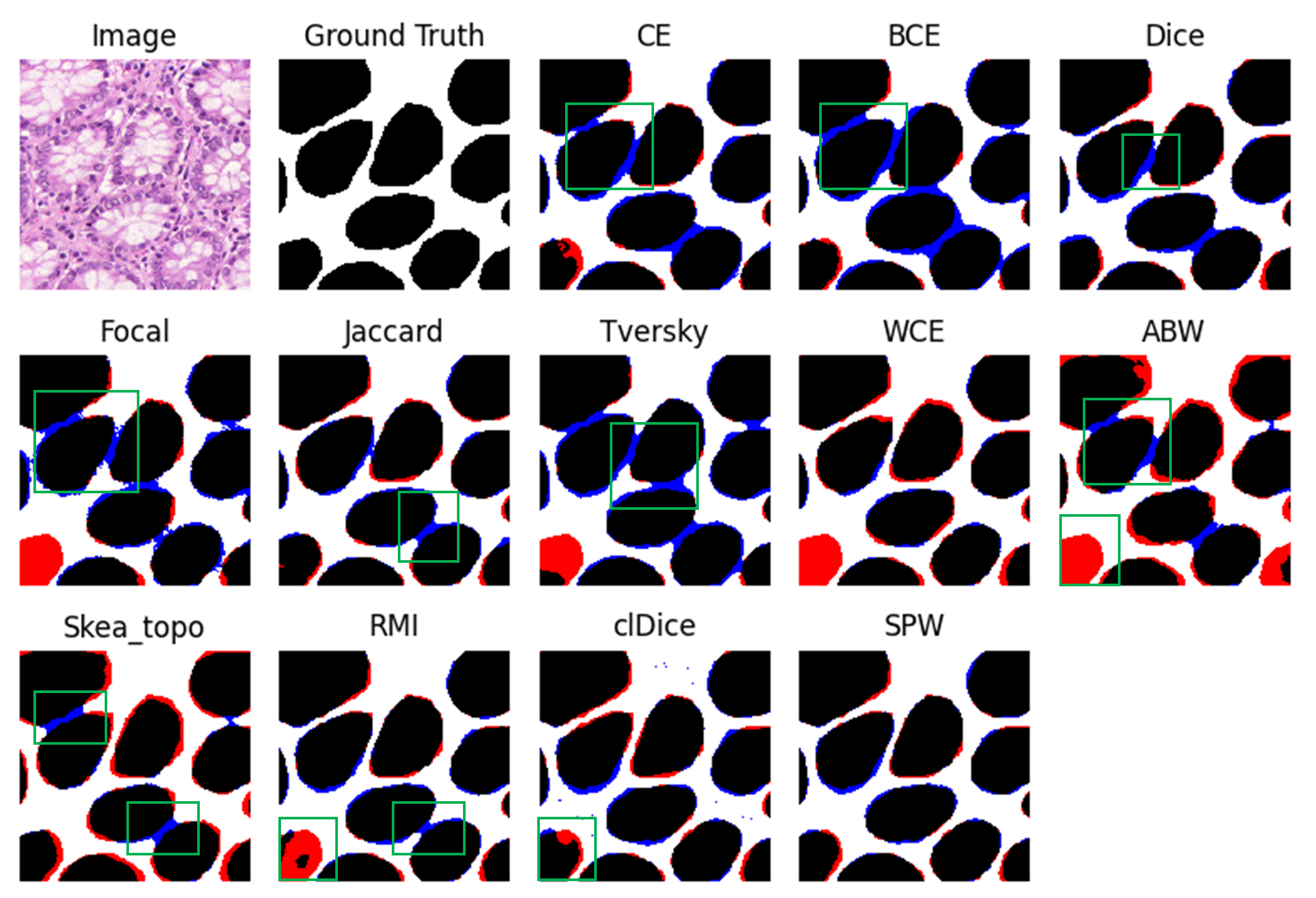}
    \caption{Qualitative results of different loss functions on the GlaS dataset. \textcolor{red}{Red}: false positive regions; \textcolor{blue}{Blue}: false negative regions. \textcolor{ForestGreen}{Green box}: challenging segmentation errors that are successfully addressed by SPW loss.}
    \label{fig:3}
\end{figure}

\begin{figure}[!h]
    \centering
    \includegraphics[width=\linewidth]{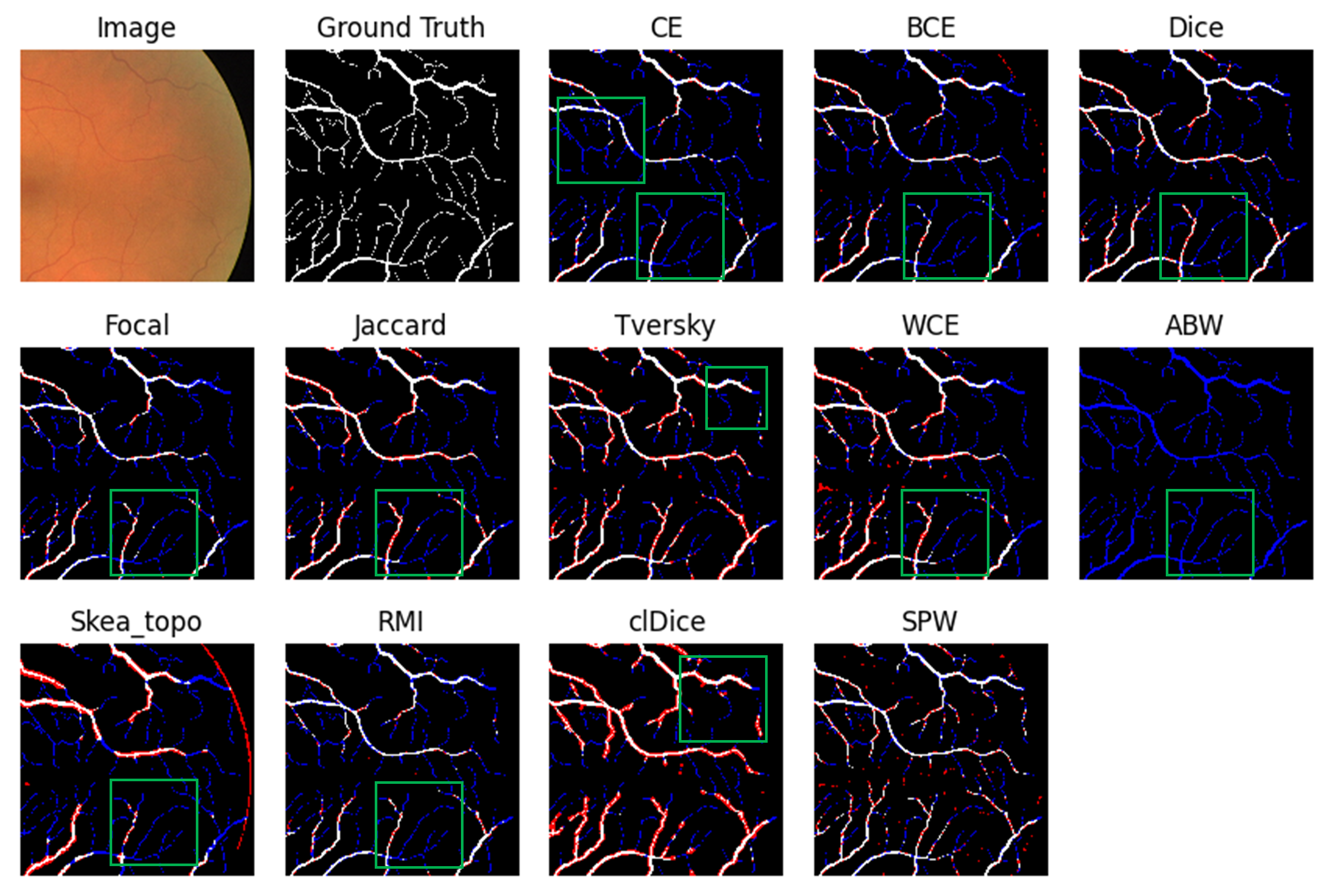}
    \caption{Qualitative results of different loss functions on the DRIVE dataset. \textcolor{red}{Red}: false positive regions; \textcolor{blue}{Blue}: false negative regions. \textcolor{ForestGreen}{Green box}: challenging segmentation errors that are successfully addressed by SPW loss.}
    \label{fig:4}
\end{figure}

As shown in Table \ref{tab:1}, the proposed SPW loss outperforms other loss functions in all metrics for all datasets, including both pixel wise metrics (mIoU and mDice) and cluster based metrics (VI and ARI). Qualitative results from SNEMI3D (Figure \ref{fig:2}), GlaS (Figure \ref{fig:3}), and DRIVE (Figure \ref{fig:4}) further demonstrate the effectiveness of SPW loss in addressing challenging segmentation issues that remain unresolved by other state-of-the-art loss functions.


\subsection{Ablation studies}
In this section, all ablation experiments were performed on the SNEMI3D dataset.

\paragraph{Hyperparameters \(\lambda\) and \(\beta\)} 
SPW loss is controlled by two key hyperparameters, \(\lambda\) and \(\beta\), which regulate the balance between different loss components and feature scales. We conduct ablation experiments to analyze their effects on performance metrics, as shown in Table \ref{fig:2}.

\textbf{Effect of \(\lambda\):}  
The parameter \(\lambda\) determines the relative influence of SPW loss in conjunction with cross-entropy loss. A very small \(\lambda\) (e.g., \(\lambda=1\)) weakens the SPW effect, while an excessively large \(\lambda\) (e.g., \(\lambda=100\)) leads to a minor performance drop, as it may overly suppress low-frequency regions. As shown in Table \ref{fig:2} (top), the best balance is achieved at \(\lambda=10\), yielding the highest mIoU (0.770) and ARI (0.629), indicating optimal segmentation performance. Thus, we use \(\lambda=10\) as the default in our main experiments (Table \ref{tab:1}).

\textbf{Effect of \(\beta\):}  
The parameter \(\beta\) controls the weighting of features across scales. When \(\beta>1\), low-frequency (large-scale) features receive higher importance, while \(\beta<1\) prioritizes high-frequency (small-scale) features. Table \ref{fig:2} (bottom) shows that \(\beta=0.9\) achieves the best trade-off, with the highest mIoU (0.770) and ARI (0.629), slightly outperforming \(\beta=1\) and \(\beta=2\). This suggests that while multi-scale features are crucial, a slight bias toward high-frequency features improves segmentation quality.

\begin{table}[h]
\footnotesize 
    \centering
    \setlength{\tabcolsep}{1pt}
    \begin{tabular}{lcccc}
        \toprule
        \multicolumn{5}{c}{\textbf{\(\lambda\) ablation test}}\\
        \hline
~  & 
mIoU$\uparrow$ & mDice$\uparrow$ & VI$\downarrow$ & ARI$\uparrow$ \\
\hline
\(\lambda=1\)   & \(0.766_{\pm0.005}\) & \(0.861_{\pm0.004}\) & \(1.21_{\pm0.07}\) & \(0.626_{\pm0.012}\) \\
\(\lambda=3\)   & \(0.765_{\pm0.004}\) & \(0.861_{\pm0.003}\) & \(1.33_{\pm0.06}\) & \(0.616_{\pm0.010}\) \\
\(\lambda=10\)  & \(\mathbf{0.770_{\pm0.003}}\) & \(\mathbf{0.864_{\pm0.002}}\) & \(1.22_{\pm0.06}\) & \(\mathbf{0.629_{\pm0.010}}\) \\
\(\lambda=30\)  & \(0.769_{\pm0.004}\) & \(0.863_{\pm0.003}\) & \(1.21_{\pm0.01}\) & \(0.622_{\pm0.008}\) \\
\(\lambda=100\) & \(0.769_{\pm0.002}\) & \(0.863_{\pm0.002}\) & \(\mathbf{1.20_{\pm0.05}}\) & \(0.620_{\pm0.011}\) \\

\toprule
        \multicolumn{5}{c}{\textbf{\(\beta\) ablation test}}\\
        \hline
~  & 
mIoU$\uparrow$ & mDice$\uparrow$ & VI$\downarrow$ & ARI$\uparrow$ \\
\hline
\(\beta=0.5\)   & \(0.764_{\pm0.005}\) & \(0.860_{\pm0.003}\) & \(1.27_{\pm0.12}\) & \(0.620_{\pm0.015}\) \\
\(\beta=0.9\)   & \(\mathbf{0.770_{\pm0.003}}\) & \(\mathbf{0.864_{\pm0.002}}\) & \(1.22_{\pm0.06}\) & \(\mathbf{0.629_{\pm0.010}}\) \\
\(\beta=1\)     & \(0.766_{\pm0.005}\) & \(0.861_{\pm0.004}\) & \(\mathbf{1.21_{\pm0.07}}\) & \(0.626_{\pm0.012}\) \\
\(\beta=2\)     & \(0.769_{\pm0.001}\) & \(0.862_{\pm0.001}\) & \(1.29_{\pm0.12}\) & \(0.616_{\pm0.014}\) \\

\toprule
        \multicolumn{5}{c}{\textbf{Prediction weight map ablation test}}\\
        \hline
~  & 
mIoU$\uparrow$ & mDice$\uparrow$ & VI$\downarrow$ & ARI$\uparrow$ \\
\hline

Label map only   & \(\mathbf{0.771_{\pm0.002}}\) & \(\mathbf{0.864_{\pm0.002}}\) & \(1.33_{\pm0.05}\) & \(0.613_{\pm0.006}\) \\
Label+pred map   & \(0.770_{\pm0.004}\) & \(0.864_{\pm0.003}\) & \(\mathbf{1.23_{\pm0.02}}\) & \(\mathbf{0.625_{\pm0.005}}\) \\

\toprule
\end{tabular}
    \caption{Results of the ablation study evaluating the impact of key hyperparameters (\(\lambda\) and \(\beta\)) and the inclusion of the prediction weight map in the SPW loss function. The \textbf{bold} numbers indicate the best performance for each metric.}
    \label{tab:2}
\end{table}

\paragraph{Inclusion of the Prediction Weight Map}  
A key novelty of our method compared to previous weighted approaches is the ability to incorporate predictions into the weight map, allowing for dynamic updates during training. To evaluate the impact of this inclusion on segmentation performance, we conducted an ablation study comparing two configurations: using only the label-derived weight map versus incorporating both the label and prediction-based weight maps.

As shown in Table \ref{tab:2} (bottom), including the prediction weight map does not significantly alter pixel-wise metrics (mIoU and mDice), but it notably improves cluster-based metrics (VI and ARI). Specifically, VI decreases from 1.33 to 1.23, while ARI increases from 0.613 to 0.625, indicating improved clustering performance. These results demonstrate that integrating the prediction into the weight map enhances segmentation quality, particularly for capturing cluster structures.

\paragraph{Computational Complexity of SPW Loss}  
The computation of SPW loss involves the following steps:  
(1) Steerable pyramid decomposition,  
(2) Envelope calculation using the analytic signal,  
(3) Upsampling with zero-padding in the Fourier domain, and  
(4) Summation across all subbands.  

For an input image of size \( H \times W \) with \( K \) orientation decompositions, the steerable pyramid decomposition has a time complexity of \(O(HW K \log(HW))\). Steps (2) and (3) can both be efficiently performed in the Fourier domain with a complexity of \(O(HW K)\), followed by an inverse Fourier transform with \(O(HW K \log(HW))\). The final summation step has a negligible complexity of \(O(HW K)\). Overall, the computational complexity of SPW loss is \(O(HW K \log(HW))\).  

In our implementation, we set \(K=4\), reducing the complexity to \(O(HW \log(HW))\), which remains scalable for high-resolution images. As shown in Table \ref{tab:3}, SPW loss is computationally more efficient than most weight-map-based or region-based loss functions, achieving a competitive balance between performance and computational cost.

\begin{table}[h]
\footnotesize 
    \centering
    \setlength{\tabcolsep}{1pt}
    \begin{tabular}{lcc}
        \toprule
        ~ & Epoch Time (s) & \(\Delta t\) to CE (s) \\
        \hline
CE & 2.04 & 0.00 \\
BCE & 2.01 & -0.03 \\
Dice & 2.01 & -0.03 \\
Focal & 2.01 & -0.03 \\
Jaccard & 2.01 & -0.03 \\
Tversky & 2.00 & -0.04 \\
WCE & 2.16 & 0.13 \\
ABW & 2.31 & 0.28 \\
Skea-topo & 2.72 & 0.68 \\
RMI & 2.28 & 0.24 \\
clDice & 2.37 & 0.33 \\
\hline
SPW & 2.20 & 0.17 \\
        \toprule
    \end{tabular}
    \caption{Training time per epoch and relative change to CE baseline for various loss functions on U-Net model with SNEMI3D dataset.}
    \label{tab:3}
\end{table}

\section{Conclusion}
In this paper, we introduced the Steerable Pyramid Weighted (SPW) loss, a novel multi-scale loss function for semantic segmentation. Unlike traditional weight map-based approaches, SPW loss dynamically adapts to both ground truth and network predictions by leveraging steerable pyramid decomposition. This enables the loss function to emphasize challenging regions across multiple frequency bands while maintaining computational efficiency.

We evaluated SPW loss on three benchmark datasets—SNEMI3D, GlaS, and DRIVE—demonstrating its superiority over 11 state-of-the-art loss functions. Our results show that SPW loss consistently improves segmentation performance, particularly in fine structures and clustered regions, achieving the best balance between pixel-wise accuracy and clustering metrics. Furthermore, our ablation studies confirm the effectiveness of incorporating multi-scale representations and prediction-based weight maps, while computational analysis highlights the method’s scalability.

Despite these advantages, SPW loss has limitations. First, while steerable pyramids naturally extend to 3D, further validation is needed for volumetric segmentation tasks such as medical CT or MRI analysis. Second, our experiments primarily focused on biomedical datasets, and future research should assess the generalizability of SPW loss to other domains, such as natural scene segmentation or remote sensing.

Overall, SPW loss offers a principled and efficient solution for enhancing segmentation performance in complex scenarios requiring detailed structural understanding. Future work will explore further optimization strategies, including adaptive scale selection and integration with self-supervised learning frameworks to enhance generalization across diverse datasets and architectures.

{
    \small
    \bibliographystyle{ieeenat_fullname}
    \bibliography{main}

\begin{thebibliography}{30}
\providecommand{\natexlab}[1]{#1}
\providecommand{\url}[1]{\texttt{#1}}
\expandafter\ifx\csname urlstyle\endcsname\relax
  \providecommand{\doi}[1]{doi: #1}\else
  \providecommand{\doi}{doi: \begingroup \urlstyle{rm}\Url}\fi

\bibitem[Arganda-Carreras et~al.(2013)Arganda-Carreras, Seung, Vishwanathan, and Berger]{arganda2013snemi3d}
Ignacio Arganda-Carreras, H~Sebastian Seung, Ashwin Vishwanathan, and Daniel~R Berger.
\newblock Snemi3d: 3d segmentation of neurites in em images.
\newblock \emph{(No Title)}, 2013.

\bibitem[Azad et~al.(2023)Azad, Heidary, Yilmaz, H{\"u}ttemann, Karimijafarbigloo, Wu, Schmeink, and Merhof]{azad2023loss}
Reza Azad, Moein Heidary, Kadir Yilmaz, Michael H{\"u}ttemann, Sanaz Karimijafarbigloo, Yuli Wu, Anke Schmeink, and Dorit Merhof.
\newblock Loss functions in the era of semantic segmentation: A survey and outlook.
\newblock \emph{arXiv preprint arXiv:2312.05391}, 2023.

\bibitem[Bagwari et~al.(2023)Bagwari, Kumar, and Verma]{bagwari2023comprehensive}
Neha Bagwari, Sushil Kumar, and Vivek~Singh Verma.
\newblock A comprehensive review on segmentation techniques for satellite images.
\newblock \emph{Archives of Computational Methods in Engineering}, 30\penalty0 (7):\penalty0 4325--4358, 2023.

\bibitem[Bokhovkin and Burnaev(2019)]{bokhovkin2019boundary}
Alexey Bokhovkin and Evgeny Burnaev.
\newblock Boundary loss for remote sensing imagery semantic segmentation.
\newblock In \emph{International symposium on neural networks}, pages 388--401. Springer, 2019.

\bibitem[Borse et~al.(2021)Borse, Wang, Zhang, and Porikli]{borse2021inverseform}
Shubhankar Borse, Ying Wang, Yizhe Zhang, and Fatih Porikli.
\newblock Inverseform: A loss function for structured boundary-aware segmentation.
\newblock In \emph{Proceedings of the IEEE/CVF conference on computer vision and pattern recognition}, pages 5901--5911, 2021.

\bibitem[Duque-Arias et~al.(2021)Duque-Arias, Velasco-Forero, Deschaud, Goulette, Serna, Decenci{\`e}re, and Marcotegui]{duque2021power}
David Duque-Arias, Santiago Velasco-Forero, Jean-Emmanuel Deschaud, Francois Goulette, Andr{\'e}s Serna, Etienne Decenci{\`e}re, and Beatriz Marcotegui.
\newblock On power jaccard losses for semantic segmentation.
\newblock In \emph{VISAPP 2021: 16th International conference on computer vision theory and applications}, 2021.

\bibitem[Guerrero-Pena et~al.(2018)Guerrero-Pena, Fernandez, Ren, Yui, Rothenberg, and Cunha]{guerrero2018multiclass}
Fidel~A Guerrero-Pena, Pedro D~Marrero Fernandez, Tsang~Ing Ren, Mary Yui, Ellen Rothenberg, and Alexandre Cunha.
\newblock Multiclass weighted loss for instance segmentation of cluttered cells.
\newblock In \emph{2018 25th IEEE International Conference on Image Processing (ICIP)}, pages 2451--2455. IEEE, 2018.

\bibitem[Gul et~al.(2022)Gul, Khan, Bibi, Khandakar, Ayari, and Chowdhury]{gul2022deep}
Sidra Gul, Muhammad~Salman Khan, Asima Bibi, Amith Khandakar, Mohamed~Arselene Ayari, and Muhammad~EH Chowdhury.
\newblock Deep learning techniques for liver and liver tumor segmentation: A review.
\newblock \emph{Computers in Biology and Medicine}, 147:\penalty0 105620, 2022.

\bibitem[Kabir et~al.(2025)Kabir, Jim, and Istenes]{kabir2025terrain}
Md~Mohsin Kabir, Jamin~Rahman Jim, and Zolt{\'a}n Istenes.
\newblock Terrain detection and segmentation for autonomous vehicle navigation: A state-of-the-art systematic review.
\newblock \emph{Information Fusion}, 113:\penalty0 102644, 2025.

\bibitem[Lin et~al.(2017)Lin, Goyal, Girshick, He, and Doll{\'a}r]{lin2017focal}
Tsung-Yi Lin, Priya Goyal, Ross Girshick, Kaiming He, and Piotr Doll{\'a}r.
\newblock Focal loss for dense object detection.
\newblock In \emph{Proceedings of the IEEE international conference on computer vision}, pages 2980--2988, 2017.

\bibitem[Liu et~al.(2024)Liu, Ma, Ban, Xie, Wang, Xue, Ma, and Xu]{liu2024enhancing}
Chuni Liu, Boyuan Ma, Xiaojuan Ban, Yujie Xie, Hao Wang, Weihua Xue, Jingchao Ma, and Ke Xu.
\newblock Enhancing boundary segmentation for topological accuracy with skeleton-based methods.
\newblock \emph{arXiv preprint arXiv:2404.18539}, 2024.

\bibitem[Liu et~al.(2022)Liu, Chen, Liu, Ban, Ma, Wang, Xue, and Guo]{liu2022boundary}
Wei Liu, Jiahao Chen, Chuni Liu, Xiaojuan Ban, Boyuan Ma, Hao Wang, Weihua Xue, and Yu Guo.
\newblock Boundary learning by using weighted propagation in convolution network.
\newblock \emph{Journal of Computational Science}, 62:\penalty0 101709, 2022.

\bibitem[Long et~al.(2015)Long, Shelhamer, and Darrell]{long2015fully}
Jonathan Long, Evan Shelhamer, and Trevor Darrell.
\newblock Fully convolutional networks for semantic segmentation.
\newblock In \emph{Proceedings of the IEEE conference on computer vision and pattern recognition}, pages 3431--3440, 2015.

\bibitem[Lu(2025)]{lu2025complex}
Renhao Lu.
\newblock Complex wavelet mutual information loss: A multi-scale loss function for semantic segmentation.
\newblock \emph{arXiv preprint arXiv:2502.00563}, 2025.

\bibitem[Milletari et~al.(2016)Milletari, Navab, and Ahmadi]{milletari2016v}
Fausto Milletari, Nassir Navab, and Seyed-Ahmad Ahmadi.
\newblock V-net: Fully convolutional neural networks for volumetric medical image segmentation.
\newblock In \emph{2016 fourth international conference on 3D vision (3DV)}, pages 565--571. Ieee, 2016.

\bibitem[Nunez-Iglesias et~al.(2013)Nunez-Iglesias, Kennedy, Parag, Shi, and Chklovskii]{nunez2013machine}
Juan Nunez-Iglesias, Ryan Kennedy, Toufiq Parag, Jianbo Shi, and Dmitri~B Chklovskii.
\newblock Machine learning of hierarchical clustering to segment 2d and 3d images.
\newblock \emph{PloS one}, 8\penalty0 (8):\penalty0 e71715, 2013.

\bibitem[Portilla and Simoncelli(2000)]{portilla2000parametric}
Javier Portilla and Eero~P Simoncelli.
\newblock A parametric texture model based on joint statistics of complex wavelet coefficients.
\newblock \emph{International journal of computer vision}, 40:\penalty0 49--70, 2000.

\bibitem[Rahman and Wang(2016)]{rahman2016optimizing}
Md~Atiqur Rahman and Yang Wang.
\newblock Optimizing intersection-over-union in deep neural networks for image segmentation.
\newblock In \emph{International symposium on visual computing}, pages 234--244. Springer, 2016.

\bibitem[Ronneberger et~al.(2015)Ronneberger, Fischer, and Brox]{ronneberger2015u}
Olaf Ronneberger, Philipp Fischer, and Thomas Brox.
\newblock U-net: Convolutional networks for biomedical image segmentation.
\newblock In \emph{Medical image computing and computer-assisted intervention--MICCAI 2015: 18th international conference, Munich, Germany, October 5-9, 2015, proceedings, part III 18}, pages 234--241. Springer, 2015.

\bibitem[Ross and Doll{\'a}r(2017)]{ross2017focal}
T-YLPG Ross and GKHP Doll{\'a}r.
\newblock Focal loss for dense object detection.
\newblock In \emph{proceedings of the IEEE conference on computer vision and pattern recognition}, pages 2980--2988, 2017.

\bibitem[Salehi et~al.(2017)Salehi, Erdogmus, and Gholipour]{salehi2017tversky}
Seyed Sadegh~Mohseni Salehi, Deniz Erdogmus, and Ali Gholipour.
\newblock Tversky loss function for image segmentation using 3d fully convolutional deep networks.
\newblock In \emph{International workshop on machine learning in medical imaging}, pages 379--387. Springer, 2017.

\bibitem[Shit et~al.(2021)Shit, Paetzold, Sekuboyina, Ezhov, Unger, Zhylka, Pluim, Bauer, and Menze]{shit2021cldice}
Suprosanna Shit, Johannes~C Paetzold, Anjany Sekuboyina, Ivan Ezhov, Alexander Unger, Andrey Zhylka, Josien~PW Pluim, Ulrich Bauer, and Bjoern~H Menze.
\newblock cldice-a novel topology-preserving loss function for tubular structure segmentation.
\newblock In \emph{Proceedings of the IEEE/CVF conference on computer vision and pattern recognition}, pages 16560--16569, 2021.

\bibitem[Simoncelli and Freeman(1995)]{simoncelli1995steerable}
Eero~P Simoncelli and William~T Freeman.
\newblock The steerable pyramid: A flexible architecture for multi-scale derivative computation.
\newblock In \emph{Proceedings., international conference on image processing}, pages 444--447. IEEE, 1995.

\bibitem[Simoncelli et~al.(1992)Simoncelli, Freeman, Adelson, and Heeger]{simoncelli1992shiftable}
Eero~P Simoncelli, William~T Freeman, Edward~H Adelson, and David~J Heeger.
\newblock Shiftable multiscale transforms.
\newblock \emph{IEEE transactions on Information Theory}, 38\penalty0 (2):\penalty0 587--607, 1992.

\bibitem[Sirinukunwattana et~al.(2017)Sirinukunwattana, Pluim, Chen, Qi, Heng, Guo, Wang, Matuszewski, Bruni, Sanchez, et~al.]{sirinukunwattana2017gland}
Korsuk Sirinukunwattana, Josien~PW Pluim, Hao Chen, Xiaojuan Qi, Pheng-Ann Heng, Yun~Bo Guo, Li~Yang Wang, Bogdan~J Matuszewski, Elia Bruni, Urko Sanchez, et~al.
\newblock Gland segmentation in colon histology images: The glas challenge contest.
\newblock \emph{Medical image analysis}, 35:\penalty0 489--502, 2017.

\bibitem[Staal et~al.(2004)Staal, Abr{\`a}moff, Niemeijer, Viergever, and Van~Ginneken]{staal2004ridge}
Joes Staal, Michael~D Abr{\`a}moff, Meindert Niemeijer, Max~A Viergever, and Bram Van~Ginneken.
\newblock Ridge-based vessel segmentation in color images of the retina.
\newblock \emph{IEEE transactions on medical imaging}, 23\penalty0 (4):\penalty0 501--509, 2004.

\bibitem[Ulrich(2006)]{ulrich2006envelope}
Timothy Ulrich.
\newblock Envelope calculation from the hilbert transform.
\newblock \emph{Los Alamos Nat. Lab., Los Alamos, NM, USA, Tech. Rep}, 2006.

\bibitem[Vinh et~al.(2009)Vinh, Epps, and Bailey]{vinh2009information}
Nguyen~Xuan Vinh, Julien Epps, and James Bailey.
\newblock Information theoretic measures for clusterings comparison: is a correction for chance necessary?
\newblock In \emph{Proceedings of the 26th annual international conference on machine learning}, pages 1073--1080, 2009.

\bibitem[Zhao et~al.(2019)Zhao, Wang, Yang, and Cai]{zhao2019region}
Shuai Zhao, Yang Wang, Zheng Yang, and Deng Cai.
\newblock Region mutual information loss for semantic segmentation.
\newblock \emph{Advances in Neural Information Processing Systems}, 32, 2019.

\bibitem[Zhu et~al.(2022)Zhu, Yin, and Meijering]{zhu2022compound}
Yanming Zhu, Xuefei Yin, and Erik Meijering.
\newblock A compound loss function with shape aware weight map for microscopy cell segmentation.
\newblock \emph{IEEE Transactions on Medical Imaging}, 42\penalty0 (5):\penalty0 1278--1288, 2022.

\end{thebibliography}
}

\end{document}